\makeatletter
\let\chemtriagekernelxhline\@xhline
\makeatother
\documentclass[12pt]{iopart}
\ifdefined\pdfminorversion\pdfminorversion=7\fi
\usepackage[T1]{fontenc}
\usepackage[utf8]{inputenc}
\expandafter\let\csname equation*\endcsname\relax
\expandafter\let\csname endequation*\endcsname\relax
\usepackage{amsmath,amssymb}
\makeatletter
\let\@xhline\chemtriagekernelxhline
\makeatother
\usepackage{lmodern,graphicx,booktabs,array,tabularx,longtable,placeins,microtype,needspace}
\usepackage{xcolor}
\usepackage[hidelinks]{hyperref}
\newcolumntype{L}[1]{>{\raggedright\arraybackslash}p{#1}}
\hypersetup{pdftitle={ChemTriage: Pre-Execution Feasibility and Safety Review of Chemical Synthesis Plans with Large Language Models},pdfauthor={Zongmin Zhang, Kexiang Hao, Xinyi Liu, Lixue Cheng}}

\makeatletter
\newcommand*{\chemtriageextlabel}[2]{%
  \@ifundefined{r@#1}{%
    \global\@namedef{r@#1}{{#2}{}{}{}{}}%
    \global\@namedef{chemtriagefallback@#1}{1}%
  }{}%
}
\AtBeginDocument{%
  \chemtriageextlabel{tab:e1}{S1}%
  \chemtriageextlabel{tab:e2}{S2}%
  \chemtriageextlabel{tab:e3}{S3}%
  \chemtriageextlabel{tab:e4}{S4}%
  \chemtriageextlabel{tab:e5}{S5}%
  \chemtriageextlabel{tab:e6}{S6}%
  \chemtriageextlabel{tab:e7}{S7}%
  \chemtriageextlabel{tab:e8}{S8}%
  \chemtriageextlabel{tab:e9}{S9}%
  \chemtriageextlabel{tab:e10}{S10}%
  \chemtriageextlabel{fig:s1}{S1}%
  \chemtriageextlabel{fig:s2}{S2}%
  \chemtriageextlabel{fig:s3}{S3}%
  \chemtriageextlabel{fig:s4}{S4}%
}
\AtBeginDocument{%
  \NewCommandCopy\chemtriageoriginalref\ref
  \RenewDocumentCommand{\ref}{s m}{%
    \IfBooleanTF{#1}{\chemtriageoriginalref*{#2}}{%
      \@ifundefined{chemtriagefallback@#2}{\chemtriageoriginalref{#2}}{\chemtriageoriginalref*{#2}}%
    }%
  }%
}
\makeatother

\begin{document}
\pagestyle{plain}
\ifdefined\XeTeXversion\fi
\title{ChemTriage: Pre-Execution Feasibility and Safety Review of Chemical Synthesis Plans with Large Language Models}
\thispagestyle{plain}
\author{Zongmin~Zhang$^{1}$, Kexiang~Hao$^{2}$, Xinyi~Liu$^2$ and~Lixue~Cheng$^{2,3,4,*}$}
\address{$^1$Department of Computer Science and Engineering, The Hong Kong University of Science and Technology, Kowloon, Hong Kong SAR 999077, China}
\address{$^2$Department of Chemistry, The Hong Kong University of Science and Technology, Kowloon, Hong Kong SAR 999077, China}
\address{$^3$IAS Center for AI for Scientific Discoveries, The Hong Kong University of Science and Technology, Kowloon, Hong Kong SAR 999077, China}
\address{$^4$Department of Chemical and Biological Engineering, The Hong Kong University of Science and Technology, Kowloon, Hong Kong SAR 999077, China}
\address{$^*$Corresponding author.}
\ead{lixuecheng@ust.hk}

\begin{abstract}
Applying a published chemical synthesis in an automated laboratory requires pre-execution review of whether current reagents, apparatus and controls support the proposed run. We evaluated how large language models convert local laboratory records into pre-execution decisions across 102 human-reviewed matched cases from 34 open-access publications, varying one decisive record so that a fixed proposal was supported, unresolved or contradicted. We compared a single-model reviewer (GPT-5.6 Sol), a specialist committee (ChemConsensus) and ChemTriage, a requirement-ledger workflow combining monotone early stopping with selective verification. The two baselines failed on opposite evidence states: across 82 shared cases, ChemConsensus approved 13 supported plans that Sol held, whereas Sol escalated 18 unresolved records that ChemConsensus approved or rejected. Youden's $J$ showed that Sol's escalations barely separated supported from unresolved plans ($+3.6$ percentage points, pp), whereas ChemConsensus escalated supported plans more often ($-16.7$~pp). Offline replay of a conflicting selenation trace showed that the committee's fail-open integration rule converted an arbiter rejection into approval. ChemTriage raised balanced three-state accuracy by 11.5~pp over Sol (70.8\% versus 59.4\%, 96 shared cases) and 27.2~pp over ChemConsensus (73.1\% versus 45.9\%, 84 shared cases), cutting estimated token cost by 23.3\% and 91.4\% across all 102 requests. It was the only workflow whose escalation-discrimination interval excluded zero ($+23.1$~pp), cutting unsupported approvals by 21.8~pp versus ChemConsensus and unnecessary holds by 21.9~pp versus Sol, with zero false rejections (0/67). Isolated quantitative checks reached 93.8--98.4\% accuracy, showing that remaining holds on supported plans reflect whole-protocol contextual judgement rather than misread quantities.
\end{abstract}


\section{Introduction}
\label{sec:introduction}

Applying a published synthesis in an automated laboratory requires evidence that the conditions of the proposed run can be met. Published procedures can already be translated into executable instructions for robotic platforms~\cite{mehr2020}, while self-driving laboratories couple experimental planning directly to physical hardware~\cite{Burger2020mobile,Tom2024sdl} and language-model agents orchestrate chemistry tools and workflows~\cite{MBran2024augmenting,Boiko2023autonomous}. Yet physical synthesis carries immediate material stakes: degraded reagents, mismatched stoichiometry, inadequate cooling or missing hazard controls can cause failed runs, equipment damage or severe accidents~\cite{Leong2025steering}. Pre-execution review must therefore inspect local evidence on the current reagent batch, available apparatus and planned controls, providing an auditable safety and readiness gate between automated planning and physical execution.

Gates of this kind are well established in safety-critical engineering and software agent security, where systems-theoretic safety controls, runtime monitors, programmable guardrails and input--control separation detect or block unsafe transitions before execution, and emulated sandboxes probe risks prior to deployment~\cite{Leveson2004accident,Leucker2009runtime,Rebedea2023nemo,Ruan2024toolemu,Xiang2025guardagent,Debenedetti2025camel}. Wet-lab chemistry poses a fundamentally harder boundary: unlike software actions checked against formal permissions or rolled back in a sandbox, physical reactions cannot be undone, and reviewers must judge heterogeneous local laboratory records against a published synthesis procedure.

\begin{figure}[tbp]
\centering
\includegraphics[width=\linewidth,height=0.70\textheight,keepaspectratio]{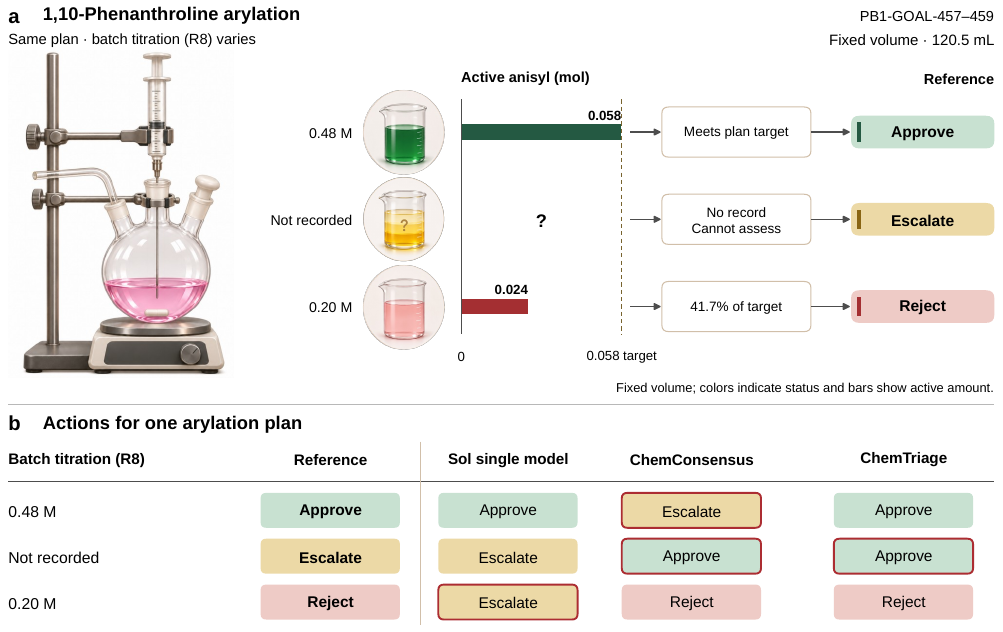}
\caption{Batch titre changes the disposition of a fixed arylation plan.
\textbf{(a)} All three records specify the same 120.5~mL transfer of dianisylmagnesium, with titre expressed on an active-anisyl basis (mol\,L$^{-1}$, M). A titre of 0.48~M supplies 0.058~mol, a missing titre leaves the active amount unknown, and 0.20~M supplies 0.024~mol (41.7\% of target), yielding reference actions of approval, escalation and rejection, respectively. Vessels depict the fixed volume schematically; bar lengths show active amount and colours indicate evidence state.
\textbf{(b)} Observed actions across the single-model reviewer (Sol), specialist committee (ChemConsensus) and requirement-ledger workflow (ChemTriage), with red outlines marking disagreements with the reference. ChemTriage approved the missing-titre record after its initial screening stage failed and fell back to a full requirement-ledger Sol review.
}
\label{fig:1}
\end{figure}

To see why this judgement is chemically subtle, consider a fixed 120.5~mL syringe transfer of dianisylmagnesium in the arylation of 1,10-phenanthroline~\cite{srch518}. At the specified batch titre of 0.48~M (on an active-anisyl basis), that fixed volume delivers 0.058~mol of active anisyl groups and satisfies the stoichiometric target. If a batch certificate omits the titre, however, the delivered amount remains unknown; at 0.20~M, the same locked transfer volume delivers only 0.024~mol (41.7\% of target). Although the reaction and proposed volume are identical across all three cases, the local records mandate three distinct pre-execution actions (Figure~\ref{fig:1}(a)): approve when decisive conditions are supported, escalate to obtain a missing fact when a condition is unresolved, or reject a fixed proposal that conflicts with documented conditions. Whereas selective prediction, deferral and abstention ask whether a model should act or abstain~\cite{pmlr-v97-geifman19a,pmlr-v119-mozannar20b,Wen2025abstention}, pre-execution chemical review must separate an unresolved information gap from a confirmed incompatibility.

In practice, a synthesis rarely hinges on one titre alone. The checks governing whether a run may proceed span multiple disciplines and experimental stages: cooling capacity and vessel records must be established before a run, whereas product characterisation verifies an outcome only after synthesis. A single model can evaluate interdependent material, equipment and procedural conditions jointly against the experimental goal, whereas a multi-agent committee can assign specialist roles to distinct hazard domains and integrate their findings into one verdict. Prior studies of multi-agent debate show that distributing reasoning can surface diverse perspectives, yet final reliability depends strongly on how conflicting judgements and confidence levels are reconciled~\cite{pmlr-v235-du24e,Liang2024mad,Chen2024reconcile,Zhu2026demystifying}. Whether single-model review or a specialist committee handles pre-execution chemical verification more reliably remains an open question.

Existing evaluations of language models in chemistry cannot settle this question, because they primarily test static domain knowledge and reasoning~\cite{Guo2023chemllmbench,Mirza2025chembench}, general laboratory and scientific safety risks~\cite{chemsafetybench,scirisk,Zhou2026benchmarking} or agent-based guardrails against dual-use chemical misuse and jailbreaks~\cite{he_ai}. None directly measures whether a reviewer adapts its disposition of a fixed synthesis proposal when a single local batch record establishes, omits or contradicts a required condition. Evaluating final verdicts alone also cannot reveal where a review breaks down: an error may arise during evidence interpretation (misreading a record or conflating post-synthesis characterisation with a pre-execution prerequisite) or during decision integration (losing a specialist's stopping judgement inside a downstream program rule). These two failure points suggest what a more reliable reviewer should do. Tying every requirement explicitly to the local records that support it would make misread or misplaced evidence visible and checkable, and preventing any later stage from overturning a documented failure would keep stopping judgements from being lost during integration.

To compare single-model and committee review directly and to test these design principles, we constructed 102 human-reviewed matched records from 34 open-access publications, holding each synthesis proposal fixed while varying one decisive local record across supported, unresolved and conflicting states. The three workflows are a single-model reviewer (GPT-5.6 Sol), a heterogeneous specialist committee (ChemConsensus) and ChemTriage, a requirement-ledger workflow that implements both principles and calls a second, stronger reviewer only when its first screen finds no failure on an explicit criterion or cannot complete. We show that the single-model and committee baselines fail on opposite evidence states, that a fail-open integration rule can erase valid stopping verdicts, and that ChemTriage, by addressing both failure points, raises balanced three-state accuracy by 11.5--27.2 percentage points (pp) over the two baselines at lower estimated cost. These results indicate that reliable pre-execution review depends less on adding reviewers than on making the evidence for each requirement explicit and preserving decisive stopping findings. {}
\section{Methods}
\label{sec:methods}

\subsection{Pre-execution review task and matched-record benchmark}

Each case captures the pre-run decision facing an automated laboratory: whether a fixed synthesis plan can proceed under the documented materials, equipment and controls. Each packet contains an experimental goal, explicit success criteria, fixed operating constraints, claims from a published procedure and local laboratory records for the proposed run. Whereas published claims specify what the synthesis requires, local records detail current reagent batches, apparatus, calibration or titration status and safety controls. The reviewer must therefore determine whether those local resources satisfy the published requirements.

Each review returns one of three actions, scored against a human-verified reference. The reference is \emph{approve} when local records support every decisive pre-execution requirement, and \emph{escalate} when no conflict is documented yet an unresolved decisive fact requires human clarification. By contrast, any documented local condition conflicting with the fixed proposal yields \emph{reject}. A confirmed conflict takes precedence over a missing fact because additional information cannot resolve an incompatible condition. Where available, retained intermediate reports distinguish record-interpretation errors from judgements lost during decision integration.

To compare these outcomes directly, the benchmark uses matched triplets in which each decision shift stems from a single local record. Each evaluation publication yields three records sharing the goal, success criteria, constraints, literature claims and surrounding context. Within a triplet, only one decisive local record varies, leaving the required condition supported, unresolved or contradicted. Source procedures were drawn from open-access Europe PMC full text~\cite{Ferguson2021europepmc} under CC BY licences and screened automatically for experimental sections rich in unit-bearing quantities and explicit hazard statements. Manual review then confirmed procedural feasibility and verified that one local record could anchor each matched comparison. Candidate local-record descriptions were drafted with large language model assistance under human chemistry supervision (Supplementary Note~S9).

The 34 evaluation publications yield 102 records, 34 per reference action, spanning reagent quantity and concentration, vessel ratings, inert-atmosphere qualification and low-temperature conditioning. Pre-freeze construction checks screened for other records restoring an omitted fact and confirmed that each conflicting variant directly violated the stated requirement. All records and reference actions then received human chemistry review and sign-off (Supplementary Note~S7). A separate development set (30 records from ten other publications) guided workflow design and controlled component comparisons, while the locked 102-record cohort was reserved for primary evaluation.

\subsection{Review workflows}

We compared three workflows that organise the same pre-execution review differently (Figure~\ref{fig:2}(a)). The workflows draw on three capability tiers of GPT-5.6 (from highest to lowest: Sol~\cite{OpenAI2026launch}, Terra and Luna), whose role assignments and execution settings appear in Table~\ref{tab:e1}.

\subsubsection{Single-model review (Sol).}
The single-model baseline tests whether one capable model can review the entire case unaided. Using a compact prompt in a single call, Sol returns a three-state action, a brief rationale and, upon approval, an executable plan. Reviewing the full packet in one context lets Sol jointly assess interdependent material, equipment and procedural requirements while leaving requirement-to-record links implicit. We refer to this configuration as Sol-compact when contrasting it with requirement-level controls.

\subsubsection{Specialist committee review (ChemConsensus).}
ChemConsensus adopts the opposite design, distributing checks across specialist agents for chemical, toxicological, fire-safety, equipment and related domains. A planner drafts a candidate plan and checks it using a deterministic structural validator, after which a router activates relevant specialists. Their reports and consistency checks inform arbitration and up to three rounds of consensus revision. Sol handles planning, routing, arbitration, adversarial audit and consistency checking, whereas domain specialists run on GPT-5.6 Terra and GPT-5.6 Luna.

However, a deterministic integration rule maps the grounded summary and candidate verdict to the final decision, rather than returning the arbiter's action directly. Under this rule, missing- or uncertain-evidence findings survive only when bound to a structured local resource identifier flagged as unresolved or conflicting, whereas confirmed-conflict findings require a valid published-claim or resource-status binding. Unmatched specialist stopping judgements are discarded as ungrounded, and an empty surviving set defaults to approval. Because all specialist reports and arbiter verdicts were retained, this rule could be replayed offline without new model calls.

\subsubsection{Requirement-level review with selective verification (ChemTriage).}
ChemTriage combines two design principles: making requirement-to-record support explicit and invoking frontier models only where a cheaper screen may err. In Stage~1, Terra reads the closed packet under a structured requirement-level prompt to build a requirement ledger linking each explicit success criterion, each fixed constraint and at most one implied goal obligation to relevant local records and citation identifiers. Writing $\mathcal{R} = \mathcal{R}_{\mathrm{SC}} \cup \mathcal{R}_{\mathrm{C}} \cup \mathcal{R}_{\mathrm{G}}$ for these three requirement types, Stage~1 assigns each $r_k \in \mathcal{R}$ a state $s_k^{(1)}$ from the ordered set $\mathsf{established} \prec \mathsf{unresolved} \prec \mathsf{contradicted}$, where a later state is more restrictive. Findings on $\mathcal{R}_{\mathrm{C}} \cup \mathcal{R}_{\mathrm{G}}$ are termed derived findings to distinguish them from findings on explicit success criteria $\mathcal{R}_{\mathrm{SC}}$.

Subsequent routing enforces one principle: a documented failure on an explicit criterion is never overturned by a more permissive stage. If Terra marks any explicit criterion unresolved or contradicted, ChemTriage stops and returns the Stage-1 ledger action (monotone early stop). Otherwise, or if Terra fails to return a valid ledger, Sol independently reviews the same original packet under identical instructions (Stage~2), free of anchoring to Terra's assessment. A Stage-2 failure on an explicit criterion likewise ends the review. Derived findings from both stages are merged conservatively, retaining the more restrictive state when both stages assess a requirement and keeping the reported state when only one stage flags it,
\begin{equation}
  \tilde{s}_k = \max\nolimits_{\prec}\bigl(s_k^{(1)}, s_k^{(2)}\bigr), \qquad r_k \in \mathcal{R}_{\mathrm{C}} \cup \mathcal{R}_{\mathrm{G}},
  \label{eq:merge}
\end{equation}
whereas explicit criteria take the Stage-2 state $\tilde{s}_k = s_k^{(2)}$. If Sol itself raises a derived finding, ChemTriage exits on the merged ledger.

A third call occurs only when every explicit criterion passes yet Terra alone raises a derived finding. A targeted Sol relevance call then inspects the original packet and Terra's derived findings to judge whether each blocks the run. Non-blocking findings become advisory; unlike explicit criteria, derived findings can therefore be relaxed at this step. Models perform all chemical interpretation, whereas the program checks only ledger coverage and citation identifiers before mapping final states $s_k^{\star}$ to an action. Active requirements $\mathcal{R}_{\mathrm{act}}$ exclude advisory findings and narrow to $\mathcal{R}_{\mathrm{SC}}$ when Stage~2 stops on an explicit criterion. Over this set, the returned action is
\begin{equation}
  \hat{a} =
  \begin{cases}
    \mathsf{R}, & \text{if } s_k^{\star} = \mathsf{contradicted} \text{ for some } r_k \in \mathcal{R}_{\mathrm{act}}, \\
    \mathsf{E}, & \text{otherwise, if } s_k^{\star} = \mathsf{unresolved} \text{ for some } r_k \in \mathcal{R}_{\mathrm{act}}, \\
    \mathsf{A}, & \text{otherwise},
  \end{cases}
  \label{eq:action-map}
\end{equation}
where $\mathsf{A}$, $\mathsf{E}$ and $\mathsf{R}$ denote \emph{approve}, \emph{escalate} and \emph{reject}. The numerical study employed a later temporal-scope variant whose relevance call covers derived findings from either stage and also determines whether a requirement applies before execution or only to post-synthesis product verification.

\begin{figure}[tbp]
  \centering
  \includegraphics[width=\linewidth,height=0.70\textheight,keepaspectratio]{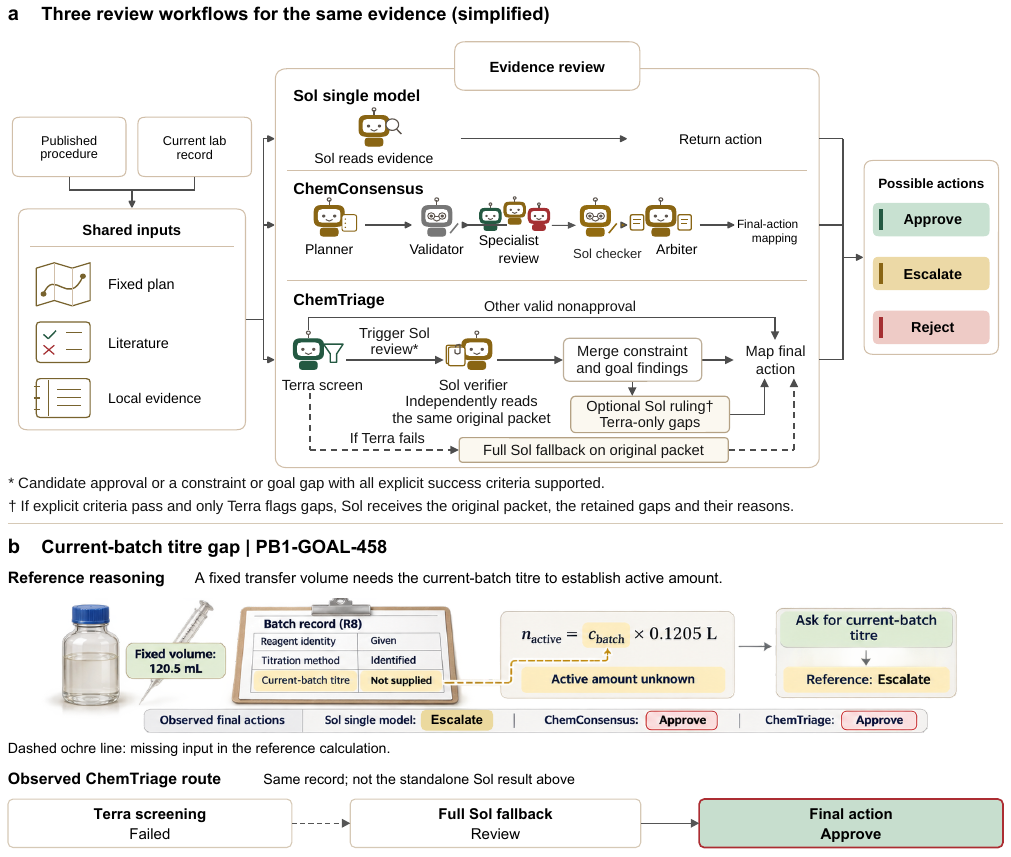}
  \caption{Review workflows and the missing-titre case.
    \textbf{(a)} Sol reviews the packet in a single call, ChemConsensus routes specialist reports through an arbiter and an integration rule, and ChemTriage begins with Terra's requirement-level screen. A failed explicit success criterion ends ChemTriage immediately (monotone early stop); otherwise, Sol independently reviews the same original packet. Derived constraint and goal findings are merged conservatively, and an optional Sol relevance call resolves findings raised only by Terra (or by either stage in the temporal-scope variant). Ochre, green and red icons denote GPT-5.6 Sol, Terra and Luna, respectively.
    \textbf{(b)} Schematic of the missing-titre case, where the dashed ochre line marks the omitted batch titre needed to calculate the active anisyl amount. The lower track traces ChemTriage's fallback to a full requirement-ledger Sol review after the Terra screen failed; red outlines mark disagreements with the reference.
  }
  \label{fig:2}
\end{figure}

\subsection{Evaluation and statistics}

Because workflow performance varies across reference classes, our primary metric is balanced three-state accuracy (BTA), the unweighted mean recall across supported, unresolved and conflicting records~\cite{Brodersen2010balanced}. To capture the safety--usability trade-off, the unsupported-approval rate (UAR) measures the fraction of returned unresolved or conflicting records approved. Conversely, the unnecessary-hold rate (UHR) measures the fraction of returned supported records escalated or rejected. Letting $N_{\hat{a}=u,\,y=v}$ denote valid returns with action $u$ and reference $v$, and $N_{y=v}$ valid returns in reference class $v$,
\begin{equation}
  \mathrm{BTA} = \frac{1}{3}\sum_{c \in \{\mathsf{A},\mathsf{E},\mathsf{R}\}} \frac{N_{\hat{a}=c,\,y=c}}{N_{y=c}}, \qquad
  \mathrm{UAR} = \frac{N_{\hat{a}=\mathsf{A},\,y\neq\mathsf{A}}}{N_{y\neq\mathsf{A}}}, \qquad
  \mathrm{UHR} = \frac{N_{\hat{a}\neq\mathsf{A},\,y=\mathsf{A}}}{N_{y=\mathsf{A}}}.
  \label{eq:metrics}
\end{equation}
Escalating nearly every case can yield high escalation recall; we therefore also report Youden's $J$~\cite{Youden1950index} on the approve-versus-escalate boundary. Escalation is treated as the positive outcome because it requests human clarification, giving
\begin{equation}
  J = P(\hat{a}=\mathsf{E} \mid y=\mathsf{E}) - P(\hat{a}=\mathsf{E} \mid y=\mathsf{A}).
  \label{eq:youden}
\end{equation}
To locate error-prone chemical domains, we grouped all 588 critical requirement rows into 11 chemistry requirement classes and compared the error rate of records carrying each class with ChemTriage's overall rate.

All metrics were computed on a primary comparison locked before reference-based scoring under a single-pass protocol (one request per record and workflow). We report accuracy over valid returns alongside coverage across all requests, computing paired differences from unrounded estimates on shared completed records. Because each publication contributes three related records, 95\% intervals are obtained from a 10,000-draw cluster bootstrap resampling whole publications with replacement~\cite{Field2007bootstrap}. Sensitivity bounds obtained by scoring all missing returns as incorrect or correct, alongside prespecified one-sided comparison criteria, are reported in Supplementary Note~S2. Resource accounting spans all requests, including partial usage before failures, at the official token rates fixed at the start of the experiment~\cite{OpenAI2026launch,OpenAI2026pricing}.

\subsection{Follow-up studies}

Five follow-up studies examined workflow mechanisms and benchmark reliability. The first three are presented in Results, whereas the last two are summarised in Results and detailed in Supplementary Notes~S4--S7.

\emph{Experimental timing.} With a recorded verifier report held fixed, one Sol call evaluated whether the flagged requirement applies before execution or only after synthesis. A temporal-scope candidate was then run fresh on the 30 development records and compared against the earlier baseline.

\emph{Integration-rule replay.} We replayed the committee's integration rule offline on a recorded trace to test whether it could discard a valid stopping judgement. All specialist reports were held fixed while each candidate action was supplied directly to the rule. A second comparison on 12 development records evaluated the original committee program against a stop-preserving variant retaining arbiter stopping judgements, with models and prompts unchanged.

\emph{Component attribution.} Four configurations were evaluated on the 30 development records to attribute ChemTriage's behaviour to its components: Sol-compact, requirement-level Sol (Sol-ledger), requirement-level Terra (Terra-ledger) and full ChemTriage. Moving from Sol-compact to Sol-ledger alters instructions, output schema and action mapping together. By contrast, the three ledger configurations share that interface, differing only in model tier and selective verification. For primary runs, initial and final ChemTriage decisions were paired to assess what Stage~2 changed.

\emph{Benchmark validity.} A provenance-phrase rule and a leave-one-publication-out naive Bayes classifier~\cite{McCallum1998nb} were trained on the changed record alone to test whether wording rather than chemistry could predict the labels, and 36 decisive descriptions from 12 publications were rewritten with all chemical facts preserved. Independently, a chemist author blindly re-reviewed the five source families behind ChemTriage's unsupported approvals; records requiring local input corrections were re-evaluated separately, and whole-publication exclusions quantified sensitivity to individual sources.

\emph{Numerical diagnostics.} Across eight publications whose decisive requirement turned on a transformable quantity, supported and conflicting records were rendered as literal values, equivalent same-unit expressions or alternate units, whereas unresolved records contrasted implicit omissions with explicit statements that the value was missing, yielding 64 records in total. Sol-ledger and the temporal-scope ChemTriage variant reviewed each record in three technical repeats, and Sol and Terra separately judged each quantity in isolation against the state inherited from its parent record. The protocol was recorded after the authors had seen the primary results for these sources but before the new calls were made. {}
\section{Results}
\label{sec:results}

This section first contrasts the failures of the single-model and committee baselines and traces their chemical and programmatic causes. It then reports ChemTriage's performance and the contribution of its components, before summarising the benchmark-validity and quantitative controls.

\subsection{Single-model and committee review succeeded under complementary evidence states}

\begin{figure}[tbp]
  \centering
  \includegraphics[width=\linewidth,height=0.70\textheight,keepaspectratio]{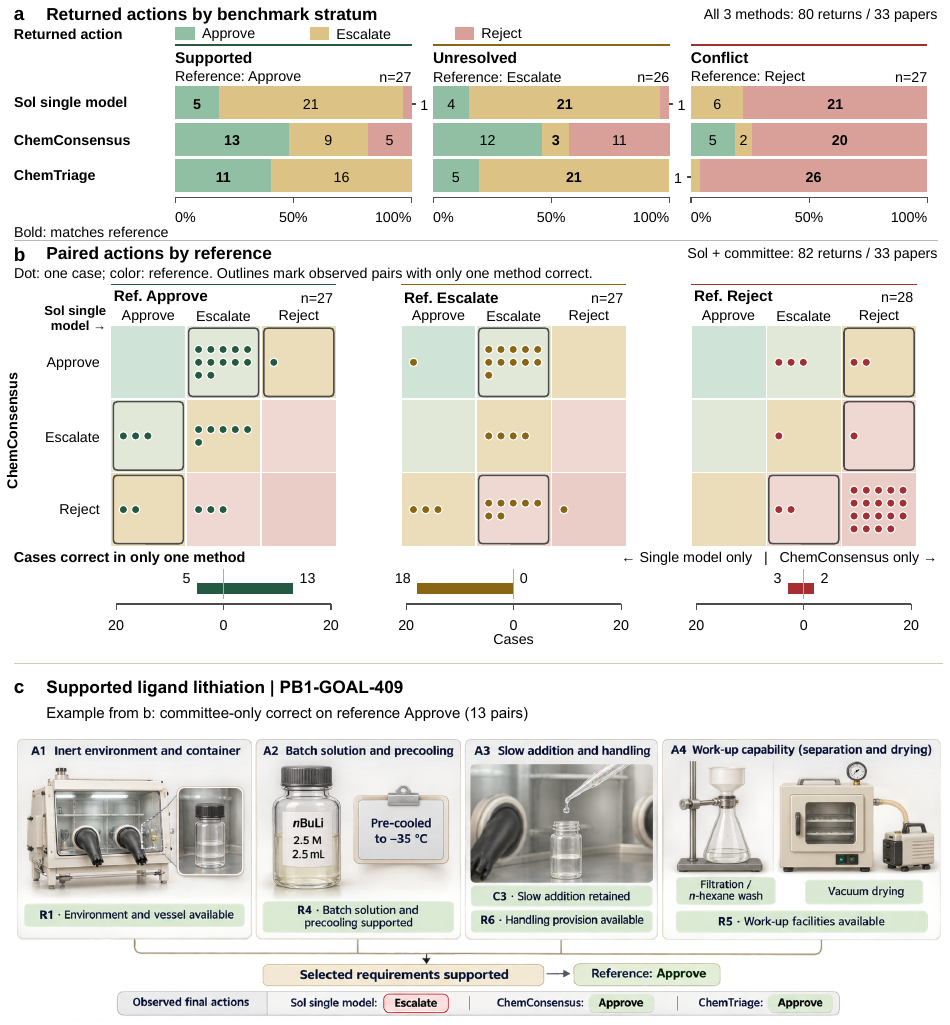}
  \caption{Review outcomes depend on which chemical conditions the records establish.
    \textbf{(a)} All three configurations returned actions for 80 shared records across 33 publications; bold counts match the reference action.
    \textbf{(b)} Sol and ChemConsensus both returned actions for 82 records across 33 publications. Each dot is one record coloured by reference action, with grey outlines marking pairs where only one method was correct. ChemConsensus approved 13 supported plans that Sol held, whereas Sol escalated 18 unresolved records that ChemConsensus approved or rejected.
    \textbf{(c)} In one of those 13 supported plans (ligand lithiation), local records documented the inert glovebox and 20~mL vial, pre-cooled 2.5~M $n$-BuLi at $-35~^\circ$C, slow-addition control and work-up facilities; ChemConsensus and ChemTriage approved whereas Sol escalated. Red outlines mark errors.
  }
  \label{fig:3}
\end{figure}

Sol and ChemConsensus failed on opposite evidence states. Of the 82 records from 33 publications that both workflows completed, Sol alone was correct on 26 and ChemConsensus alone on 15 (Figure~\ref{fig:3}(b)). The committee's exclusive successes were almost all approvals of supported proposals (13 of 15), while Sol's exclusive successes were mostly escalations of unresolved records (18 of 26). On the 27 shared unresolved records, ChemConsensus issued a definite action on 23, approving 12 and rejecting 11, and requested clarification on only four (Supplementary Note~S2). Figure~\ref{fig:3}(a) shows the outcomes of all three workflows on the 80 records completed by all three.

The same reversal occurred within a single reaction, where changing only the local batch record determined which baseline succeeded. In the organomagnesium arylation of 1,10-phenanthroline~\cite{srch518}, Sol approved the supported 0.48~M batch while ChemConsensus escalated; without a recorded titre, Sol escalated while ChemConsensus approved. When the titre fell to 0.20~M, so that the fixed 120.5~mL transfer delivered only 41.7\% of the target amount, ChemConsensus correctly rejected the proposal, but Sol escalated it (Figure~\ref{fig:1}(b)). The opposite pattern appeared in the supported ligand lithiation reported by Liao et~al.~\cite{srch405}, where the local records documented the relevant handling and work-up conditions; ChemConsensus and ChemTriage approved the proposal, while Sol held it for clarification (Figure~\ref{fig:3}(c)). Since neither baseline reliably handled both supported approvals and missing-fact escalations, the next two analyses examined failures in reading the records and integrating the findings.

\subsection{Interpretation errors involve analytical records and experimental timing}

\begin{figure}[tbp]
  \centering
  \includegraphics[width=\linewidth,height=0.70\textheight,keepaspectratio]{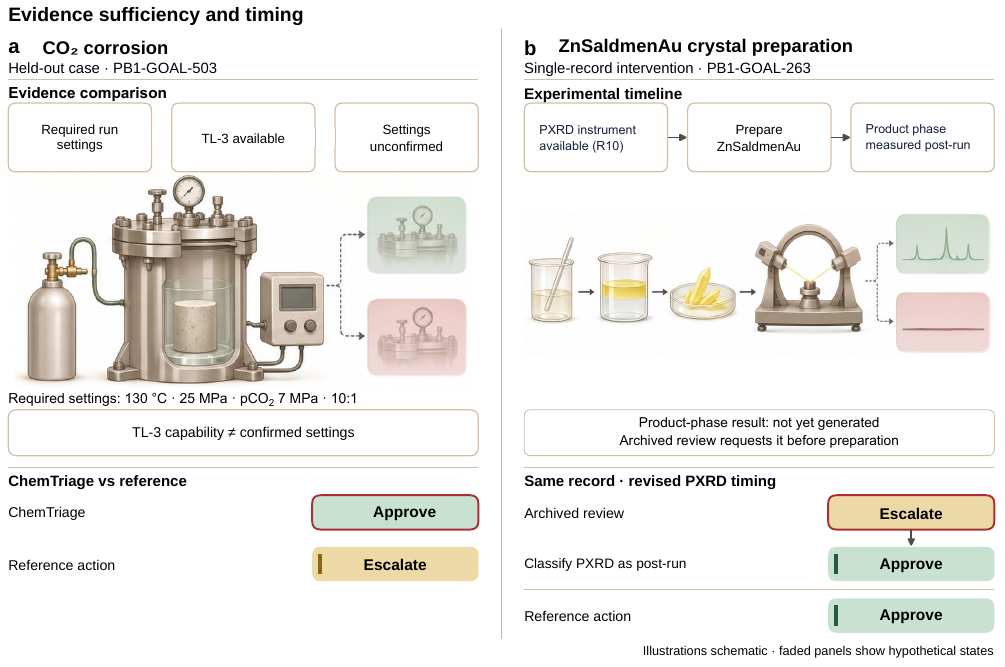}
  \caption{Equipment availability and product measurements answer different pre-execution questions.
    \textbf{(a)} For high-pressure $\mathrm{CO_2}$ cement carbonation (corrosion test), the record established TL-3 autoclave availability without confirming the required run settings; ChemTriage approved against a reference action of escalation.
    \textbf{(b)} For ZnSaldmenAu preparation (development cohort), powder X-ray diffraction (PXRD) was available, but product phase can be measured only after synthesis. With the verifier report held fixed, one Sol temporal-scope call classified the phase check as post-synthesis and shifted escalation to the reference approval. Red outlines mark errors and faded schematics show hypothetical states.
  }
  \label{fig:4}
\end{figure}

Individual error traces and a targeted development intervention on experimental timing show where record interpretation breaks down. Unsupported approvals arose when local records named the relevant reagent or instrument without confirming the decisive run parameter. In the missing-titre arylation record~\cite{srch518}, the certificate identified dianisylmagnesium and the titration method but omitted the active batch concentration (Figure~\ref{fig:2}(b)). A high-pressure $\mathrm{CO_2}$ cement-carbonation (corrosion) procedure~\cite{srch527} showed the same gap: the published method specified the temperature, pressure and fluid-to-cement ratio, but the local record confirmed only that a high-pressure autoclave was available (Figure~\ref{fig:4}(a)). Sol escalated both cases, whereas ChemConsensus and ChemTriage approved both, with ChemTriage approving via Sol fallback after Terra failed in the arylation case and via two-stage approval in the carbonation case (Supplementary Note~S10).

Misjudged timing produced the opposite failure, also involving an analytical record. In a development synthesis of the ZnSaldmenAu Schiff-base metal complex~\cite{srcpmc12787429}, powder X-ray diffraction (PXRD) verifies the product phase only after synthesis. The local records confirmed that a PXRD instrument was available, yet the archived ChemTriage review requested product diffraction data before the run. With the verifier report held fixed, a single Sol temporal-scope call (USD~0.09) classified the phase measurement as post-synthesis and changed the decision from \emph{escalate} to \emph{approve} (Figure~\ref{fig:4}(b)). In a separate run on the 30 development records, the temporal-scope candidate and the earlier baseline both reached 96.7\% BTA (28/29 and 29/30, respectively), with one correction and one regression. The timing correction therefore fixed the targeted case without changing overall accuracy on the development records.

\subsection{Programmatic integration rules can erase valid stopping judgements}

\begin{figure}[tbp]
  \centering
  \includegraphics[width=\linewidth,height=0.70\textheight,keepaspectratio]{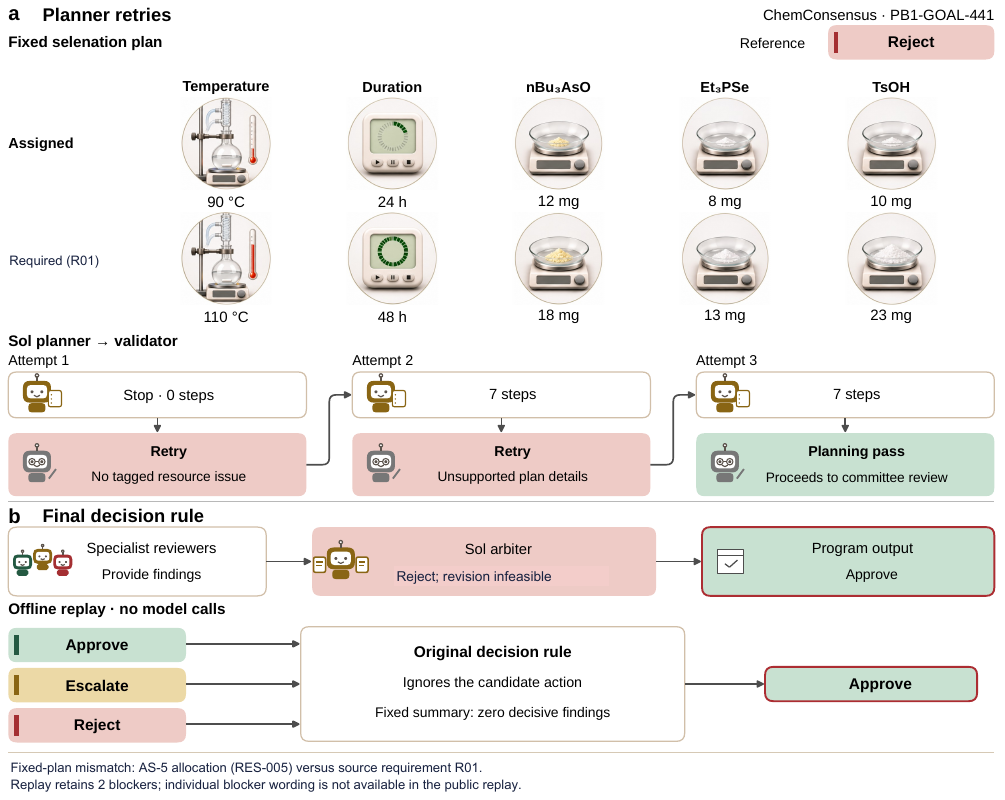}
  \caption{A recorded stopping judgement was erased by the committee's final integration rule.
    \textbf{(a)} In the post-audit corrected selenation record (PB1-GOAL-441), assigned reagent allocation (AS-5: 12/8/10~mg at 90~$^\circ$C for 24~h) conflicted with published requirements (18/13/23~mg at 110~$^\circ$C for 48~h); the planner stopped on attempt~1 before passing validation on attempt~3.
    \textbf{(b)} The arbiter returned rejection, judged revision infeasible and recorded two blockers, yet the program output approval. Zero-call offline replay with the committee reports held fixed mapped all three candidate actions (\emph{approve}, \emph{escalate}, \emph{reject}) to approval. Red outlines mark disagreements with the reference.
  }
  \label{fig:5}
\end{figure}

Chemical misinterpretation did not fully explain the committee's errors, as its integration rule could also discard a recorded stopping verdict before output. In a chalcogen-transfer selenation case corrected after the blinded human audit~\cite{srch501}, the assigned allocation specified a lower temperature, a shorter reaction time and smaller amounts of all three reagents than the published procedure required (Figure~\ref{fig:5}(a)). The Sol arbiter recommended rejection, citing an infeasible revision and two blockers. Despite this, the rule returned \emph{approve}. In a zero-call offline replay with all specialist reports and summaries held fixed, all three candidate actions (\emph{approve}, \emph{escalate} and \emph{reject}) were supplied to the original rule, and each mapped to \emph{approve} (Figure~\ref{fig:5}(b)). This outcome follows directly from the rule described in Methods: specialist blockers failing the required grounding criteria were discarded, and the resulting empty decisive set overrode the arbiter's candidate action and defaulted to approval.

If the integration rule causes this loss, preserving the arbiter's stopping judgements should change the committee's decisions. A stop-preserving modification of ChemConsensus, tested on 12 development records from four publications with ten shared terminal returns, illustrated this effect. With models, prompts and decoding settings unchanged, retaining the arbiter's stopping judgements and blocker flags reduced unsupported approvals from 3/6 to 1/6, removed the single false rejection of an unresolved case and raised BTA from 47.2\% to 63.9\%, while unnecessary holds increased from 1/4 to 3/4 (Figure~\ref{fig:s1}, Table~\ref{tab:e4}). The integration rule therefore changes the committee's balance between unsupported approvals and holds, independently of its specialists' reasoning.

\subsection{ChemTriage raised balanced accuracy, boundary discrimination and cost efficiency}

ChemTriage was designed to address both failure points. It was evaluated on all 102 primary requests per workflow (Table~\ref{tab:main-summary}) and returned a terminal action for 100 of them, whereas ChemConsensus returned 86, mainly because of provider failures during multi-call rounds. ChemTriage also achieved the highest overall BTA (71.3\% against 59.8\% for Sol and 47.3\% for ChemConsensus), and this ranking held when every missing return was scored as either wrong or correct. No false rejections occurred among its 67 supported and unresolved returns, against 17 of 57 for ChemConsensus, and its single approval of a confirmed conflict (1 of 33, against 5 of 29 for ChemConsensus) occurred on one of the two source families later corrected during the blinded human audit~\cite{srch424}.

\begin{table}[tbp]
\caption{\textbf{Primary evaluation summary across all 102 matched requests per workflow.} Overall metrics cover each system's valid terminal returns; paired contrasts are restricted to shared returns with nominal 95\% source-publication cluster bootstrap intervals (10,000 draws; full breakdowns in Tables~\ref{tab:e2}, \ref{tab:e3} and~\ref{tab:e9}). Resources cover all 102 requests per workflow. Follow-up cohorts (development controls, wording rewrites and numerical probes) are detailed in Supplementary Notes~S3--S7.}
\label{tab:main-summary}
\centering
\footnotesize
\setlength{\tabcolsep}{3pt}
\renewcommand{\arraystretch}{1.15}
\begin{tabular}{@{}L{\dimexpr 0.33\linewidth-\tabcolsep\relax}L{\dimexpr 0.20\linewidth-2\tabcolsep\relax}L{\dimexpr 0.21\linewidth-2\tabcolsep\relax}L{\dimexpr 0.26\linewidth-\tabcolsep\relax}@{}}
\toprule
\textbf{Measure} & \textbf{Sol} & \textbf{ChemConsensus} & \textbf{ChemTriage} \\
\midrule
Terminal returns / requested & 98/102 (96.1\%) & 86/102 (84.3\%) & 100/102 (98.0\%) \\
Overall BTA & 59.8\% & 47.3\% & 71.3\% \\
Recall (A / E / R, \%) & 15.6 / 84.8 / 78.8 & 51.7 / 14.3 / 75.9 & 38.2 / 84.8 / 90.9 \\
False rejections ($\mathsf{A}+\mathsf{E}$) & 2/65 (3.1\%) & 17/57 (29.8\%) & 0/67 (0.0\%) \\
Youden's $J$, pp [95\% CI] & $+3.6$ [$-11.7$, $+19.0$] & $-16.7$ [$-36.2$, $+1.5$] & $+23.1$ [$+10.5$, $+37.9$] \\
Paired BTA (CT $-$ col.), pp [95\% CI] & $+11.5$ [$+2.6$, $+20.4$] & $+27.2$ [$+18.0$, $+35.9$] & Reference ($n=96$/$84$) \\
Paired UAR / UHR (CT $-$ col.), pp & $+3.1$ / $-21.9$ & $-21.8$ / $+10.3$ & Reference \\
Completed model calls & 98 & 1,487 & 143 \\
Estimated cost, USD & 20.99 & 186.94 & 16.09 ($-23.3$\% / $-91.4$\%) \\
Mean latency (s/request) & 160.2 & 1,203.0 & 130.4 \\
\bottomrule
\end{tabular}
\end{table}

\begin{figure}[tbp]
  \centering
  \includegraphics[width=\linewidth,height=0.70\textheight,keepaspectratio]{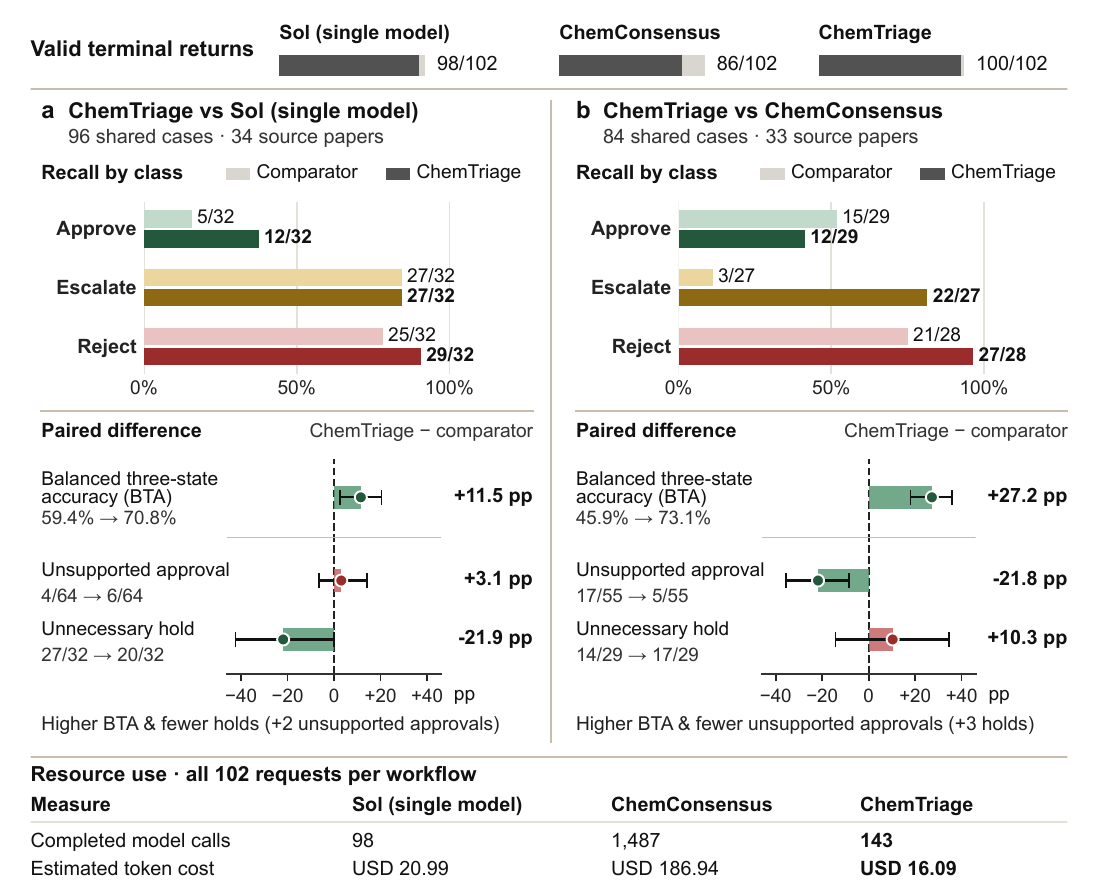}
  \caption{Primary accuracy, error trade-offs and resource use across the three review workflows.
    \textbf{(a)} Paired comparison of ChemTriage and Sol (single model) on 96 shared cases from 34 publications.
    \textbf{(b)} Paired comparison of ChemTriage and ChemConsensus on 84 shared cases from 33 publications. Top bars report per-class recall; middle forest plots show paired differences (ChemTriage minus comparator) with nominal two-sided 95\% publication-cluster bootstrap intervals (10,000 draws; green favourable, red unfavourable). Bottom table summarises completed model calls and estimated token costs across all 102 requests per workflow.
  }
  \label{fig:6}
\end{figure}

The advantage held in paired comparisons (Figure~\ref{fig:6}). ChemTriage exceeded Sol by 11.5~pp in BTA (70.8\% versus 59.4\%; 95\% CI $+2.6$ to $+20.4$~pp; 96 shared cases) and ChemConsensus by 27.2~pp (73.1\% versus 45.9\%; 95\% CI $+18.0$ to $+35.9$~pp; 84 shared cases). Most of the gain over Sol came from approving 12 of 32 supported plans against 5 for Sol, alongside rejecting four more conflicting plans while matching Sol's 27 of 32 escalations.

Identical escalation recall concealed qualitatively different behaviour, revealed by Youden's $J$ (Equation~\ref{eq:youden}, Table~\ref{tab:main-summary}). Sol escalated 81.2\% of supported plans and 84.8\% of unresolved records ($J = +3.6$~pp), a pattern indistinguishable from indiscriminate holding, whereas ChemConsensus escalated supported plans more often than unresolved ones ($J = -16.7$~pp) because it approved or rejected most unresolved records. ChemTriage was the only workflow whose interval excluded zero ($J = +23.1$~pp), exceeding Sol by 18.8~pp on 64 shared boundary cases (95\% CI $-3.1$ to $+41.5$~pp).

ChemTriage also improved the safety--usability trade-off at lower cost (Table~\ref{tab:main-summary}). Against ChemConsensus, unsupported approvals fell from 17 to 5 of 55 shared non-approve cases ($-21.8$~pp), with unnecessary holds shifting from 14 to 17 of 29 supported plans. Relative to Sol, unsupported approvals were 6 versus 4 of 64 shared non-approve cases ($+3.1$~pp; 95\% CI $-6.6$ to $+14.1$~pp), while unnecessary holds dropped by 21.9~pp; on the 19 shared primary records where both workflows erred, Sol and ChemTriage returned the same incorrect action on 18. Across all 102 requests, ChemTriage cut estimated token cost by 23.3\% below Sol and 91.4\% below ChemConsensus with the lowest mean latency.

Error rates were uniform across most chemical domains: in ten of 11 requirement classes, including stoichiometry, inert-atmosphere control and thermal envelopes, error rates remained close to the 29.0\% overall terminal error rate (29/100 valid returns), with bootstrap intervals spanning zero (Table~\ref{tab:e10}). Only records carrying the \emph{Analysis, monitoring and record} class showed a nominally elevated error rate ($+7.7$~pp; 95\% CI $+0.7$ to $+15.9$~pp), reflecting the analytical-record failures traced above.

\subsection{Requirement-ledger decomposition and selective verification play complementary roles}

\begin{figure}[tbp]
  \centering
  \includegraphics[width=\linewidth,height=0.70\textheight,keepaspectratio]{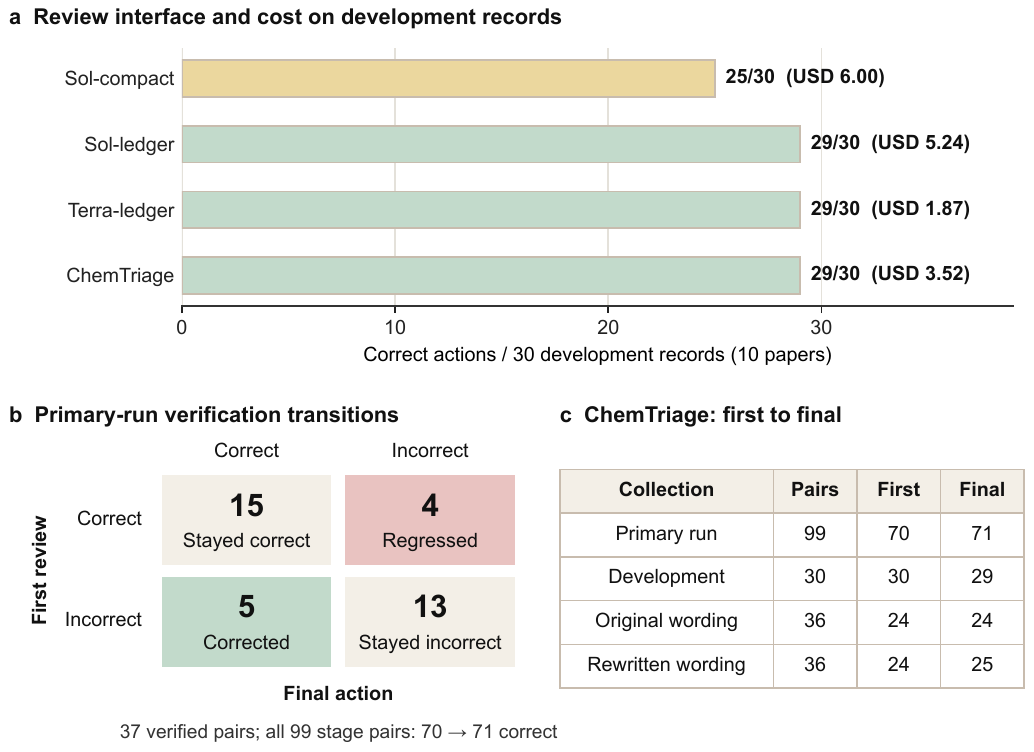}
  \caption{Requirement-level review interface and recorded stage transitions.
    \textbf{(a)} On 30 development records from ten publications, Sol-compact returned 25/30 correct actions (USD~6.00), whereas Sol-ledger (USD~5.24), Terra-ledger (USD~1.87) and ChemTriage (USD~3.52) each returned 29/30.
    \textbf{(b)} Across the 37 primary records that received additional verification and retained both stage decisions, five changed from incorrect to correct, four changed from correct to incorrect, 15 remained correct and 13 remained incorrect.
    \textbf{(c)} Initial-to-final correct counts across collections: 99 paired primary runs (70 to 71 correct out of 100 valid returns), 30 development runs (30 to 29 within ChemTriage versus 29/30 for the independently run Terra-ledger arm in (a)), and 36 original versus rewritten wording pairs reusing 12 source publications.
  }
  \label{fig:7}
\end{figure}

Because ChemTriage combines a requirement-ledger interface with two-stage verification, the contribution of each component to these gains was isolated on the 30 development records (Table~\ref{tab:e4}, Figure~\ref{fig:7}(a)). Replacing the compact Sol prompt with the ledger interface raised accuracy from 25/30 to 29/30 and eliminated both unsupported approvals. With the interface held fixed, Terra-ledger and the full ChemTriage workflow also reached 29/30 without unsupported approvals, and ChemTriage did so at an estimated cost of USD~3.52, compared with USD~5.24 for Sol-ledger (a saving of 32.9\% under identical instructions) and USD~1.87 for Terra-ledger.

Paired initial and final decisions from the primary runs show the contribution of Stage~2 on unseen publications. Of the 99 primary runs with a valid Stage-1 screen (one further case fell back directly to Sol after Terra failed), Stage~1 was correct on 70 and the final ChemTriage action on 71. Among the 37 records routed to Stage~2, five became correct, four became incorrect and 28 remained unchanged (Figure~\ref{fig:7}(b,c)). Stage~1 already produced no false rejections, and Stage~2 reduced unsupported approvals from eight to five at the cost of two additional unnecessary holds.

Grouping the 99 Stage-1-screened returns by route shows where each mechanism operates. Monotone early stopping settled 62 cases at Stage~1 (51 correct), a route that by design never approves a plan. Stage-2 review of the 23 candidate approvals intercepted three unsupported approvals while converting four supported plans into holds. Among the 14 cases with derived findings, Sol confirmed non-approval in ten, and the relevance call on the remaining four recovered two supported proposals without approving any non-approve record, reducing holds in this group from ten to eight.

Stage-2 verification and relevance calls accounted for 41.2--47.0\% of the estimated cost in the follow-up collections that logged per-stage usage. Taken together, the ledger interface accounted for the gain in accuracy on the development records, while early stopping and selective verification acted mainly on cost and on the balance between unsupported approvals and holds.

\medskip\noindent\textbf{Summary of supporting controls.} Additional validity, audit and quantitative controls are detailed in Supplementary Notes~S4--S7 (Tables~\ref{tab:e5}--\ref{tab:e8} and Figures~\ref{fig:s2}--\ref{fig:s4}). Although the oracle-isolated changed record remained lexically separable (naive Bayes 92/102 originally; 28/36 to 30/36 after rewrites), fact-preserving wording rewrites across 36 records shifted ChemTriage far less than Sol-ledger ($+2.8$ versus $+13.9$~pp), while excluding either the two audit-corrected publications~\cite{srch501,srch424} or a germanium-precursor source with inconsistent reported stoichiometry~\cite{srch515} preserved ChemTriage's paired BTA gain over Sol at $+12.2$ and $+10.8$~pp, respectively. Across 384 whole-plan numerical-perturbation reviews, neither Sol-ledger nor the temporal-scope ChemTriage variant approved any conflicting variant (0/72 per workflow; 0/24 in every representation arm), explicitly stating a missing value reduced unsupported approvals (from 5/24 to 2/24 for Sol-ledger and from 4/24 to 3/24 for ChemTriage), and 10.9\% of variants (7/64 per workflow) changed action across three temperature-0 repeats. Whereas whole-plan approval of supported numerical variants was 13.9--19.4\%, isolated quantity checks reached 93.8\% accuracy for Sol (60/64) and 98.4\% for Terra (63/64), detecting all 16/16 missing values; whole-plan traces, illustrated by the phosphorus-doped carbon ampoule preparation~\cite{srch409}, showed that remaining holds on supported plans can reflect protocol-level hazards beyond the focal quantity, such as an unverified vessel filling limit. {}
\section{Discussion}
\label{sec:discussion}

\subsection{Pre-execution review depends on judging which facts are decisive}

Our results show that neither single-model nor committee review is reliable alone because both break down at two distinct failure points: interpreting which local records govern execution, and preserving stopping judgements during integration.

The first failure point concerns contextual judgement rather than basic chemical reading. In isolated quantitative controls, both models read numerical requirements almost without error (93.8--98.4\%) and detected every missing value. Within full multi-record packets, however, reviewers must distinguish unstated conventions from execution prerequisites and pre-run readiness from post-synthesis characterisation. The missing batch titre~\cite{srch518}, unconfirmed autoclave setpoints~\cite{srch527} and pre-run PXRD demand~\cite{srcpmc12787429} all turn on what an analytical record establishes and when it applies (Figures~\ref{fig:2}(b) and~\ref{fig:4}), consistent with the elevated error rate on records carrying \emph{Analysis, monitoring and record} requirements. Even when focal quantities are verified, holds on supported plans can reflect protocol-level hazards beyond the targeted number, such as requiring the filling limit of a sealed quartz ampoule under pyrolysis overpressure~\cite{srch409}.

This requirement also distinguishes pre-execution review from selective prediction and abstention, which ask only whether to act~\cite{pmlr-v97-geifman19a,Wen2025abstention}. A laboratory gate must further separate a human-resolvable information gap from an unfixable physical conflict while still distinguishing unresolved gaps from supported runs. Across the approve-versus-escalate boundary, only ChemTriage separated supported from unresolved plans above chance (Table~\ref{tab:main-summary}). Effective clarification requests must therefore target unresolved facts that are knowable before execution and decisive for run safety or feasibility.

\subsection{Specialist review depends on both reasoning and stop-preserving integration}

The second failure point lies downstream of model reasoning. Distributing chemical, toxicological and equipment checks across specialist agents recovered supported approvals held by Sol, demonstrating the value of domain decomposition. Yet specialist reasoning produces safe outcomes only when stopping verdicts survive integration: erasing an arbiter's rejection inside a fail-open rule~\cite{srch501} (Figure~\ref{fig:5}(b)) and reducing unsupported approvals at the cost of extra holds in the stop-preserving development repair confirm that the integration rule itself sets the safety--usability operating point. This mirrors findings that auditing structured reasoning traces outperforms aggregating opaque terminal verdicts in multi-agent systems~\cite{agentauditor}.

\subsection{Design principles for requirement-level selective verification}

ChemTriage's architecture maps directly onto these two failure points. First, explicit requirement-to-record ledgers with deterministic action mapping make each decision auditable and drove most of the development accuracy gain, bringing both frontier and lower-tier models near ceiling performance (Figure~\ref{fig:7}(a)) while yielding zero false rejections on the primary cohort (Table~\ref{tab:main-summary}). Second, monotone early stopping protects documented failures on explicit criteria, while selective verification concentrates frontier-model compute on risk control. Because Terra settled most cases at Stage~1, ChemTriage cost less than either baseline on primary requests and 32.9\% less than prompt-matched Sol-ledger on development records (while Terra-ledger remained cheaper still when Stage-2 verification was omitted). On primary runs, Stage~2 functioned chiefly to intercept unsupported approvals at the cost of extra holds; across the wording-rewrite cohort, ChemTriage also shifted less than Sol-ledger ($+2.8$ versus $+13.9$~pp).

Because Stage-2 verification and relevance calls consumed 41.2--47.0\% of follow-up cost, replacing full-packet re-reads with targeted checks focused on high-error domains, including batch-titre stoichiometry, cross-record setpoint confirmation and pre- versus post-execution timing, offers a direct path to higher precision per unit of compute.

\subsection{Premature definite actions and correlated errors suggest testable mechanisms}

Although these principles address both primary failure modes, three mechanisms behind residual errors warrant direct testing. First, explicitly stating a missing value reduced unsupported approvals even though isolated checks caught every omission; tracking implicit gap recognition inside full packets would separate missed gaps from gaps recognised but overridden, extending work on selective abstention~\cite{abstentionbench}. Second, because ChemConsensus issued definite verdicts on most unresolved records it completed, and RLHF can reward verbalised overconfidence~\cite{leng_reward} while multi-agent debate can suppress minority dissent through peer sycophancy~\cite{yao_peer}, holding records and initial escalations fixed while varying peer consensus would test whether social conformity erodes clarification requests in chemical committees. Third, ChemTriage and Sol chose the same incorrect action on 18 of 19 shared primary errors, echoing correlated-error and agent-failure studies~\cite{correlatederrors,Yang2024agent,ICLR2025_1b126cc3,Cemri2025mas}; auditing cited requirement rows across those errors would test whether both workflows share an upstream misreading.

\subsection{Scope and evaluation boundaries}

Four boundaries define where these conclusions apply and which tests extend them. First, the primary benchmark covers 102 matched records from 34 publications evaluated once per workflow within one model family (GPT-5.6 Sol, Terra and Luna), whereas repeat stability was measured on the numerical cohort (10.9\% action flips across three temperature-0 repeats) and the 11-class error breakdown is exploratory. Evaluating additional model families, repeating primary passes and running single-stage Sol-ledger directly on the primary cohort would test cross-family transportability and isolate interface gains on unseen procedures. Second, the wording, numerical and blinded audit studies sampled subsets of the 34 publications; nevertheless, excluding the two audit-corrected sources or a source with inconsistent stoichiometry~\cite{srch515} preserved ChemTriage's paired BTA gain over Sol at $+12.2$ and $+10.8$~pp. Third, as shown in annotation-artifact, counterfactual-benchmark and shortcut-learning studies~\cite{Gururangan2018artifacts,Kaushik2020counterfactual,Geirhos2020shortcut}, the isolated changed record remains lexically informative even after fact-preserving rewrites; pairing full-packet lexical baselines with naturally authored laboratory batch logs would remove construction templates. Fourth, our metrics score agreement with human-verified reference actions on fixed proposals under official rate-card pricing, providing a controlled precursor to closed-loop physical execution. {}
\section{Conclusion}
\label{sec:conclusion}

Pre-execution safety review in automated chemical synthesis breaks down along two distinct axes: misjudging what current laboratory records establish for the relevant experimental stage, and silently losing valid stopping judgements inside downstream integration rules. Across 102 matched cases from 34 publications, single-model review (GPT-5.6 Sol) and specialist committee review (ChemConsensus) failed on opposite evidence states, while offline replay showed that a fail-open integration rule can convert an explicit arbiter rejection into approval.

ChemTriage addresses both failure modes by pairing an explicit requirement-to-record ledger with monotone early stopping and selective verification. Across shared cases, it raised balanced three-state accuracy by 11.5~pp over Sol and 27.2~pp over ChemConsensus; across all primary returns, it achieved the only escalation-discrimination interval above zero and zero false rejections (0/67), while cutting estimated token cost by 23.3\% and 91.4\% across all 102 requests. Development component controls confirmed that explicit requirement decomposition drove the accuracy gain while selective verification delivered cost control and risk filtering.

These findings establish two design requirements for reliable pre-execution gates in self-driving laboratories: every requirement-to-record judgement should be recorded in an auditable ledger sensitive to experimental timing, and integration logic must preserve decisive stopping findings. Extending this requirement-level framework to routine laboratory batch logs and closed-loop wet-lab synthesis is the next step towards safe, trustworthy autonomous execution. {}
\FloatBarrier
\FloatBarrier
\ack
This work was supported by start-up funding from The Hong Kong University of Science and Technology (HKUST). Z.Z. acknowledges support from the HKUST Undergraduate Research Opportunities Program (UROP).

\section*{Data and code availability}
The ChemTriage source code, evaluation scripts, matched benchmark records, reference actions, and static prompts are publicly available at \url{https://github.com/AI4QC/ChemTriage}.

\section*{Author contributions}
Z.Z. developed the ChemTriage and ChemConsensus software, ran the evaluations, performed the formal analysis, prepared the figures and wrote the original draft. K.H. designed the review workflows and evaluation methodology, constructed the matched-record benchmark and contributed to the investigation. X.L. curated and validated the benchmark records and reference actions through human chemistry review. L.C. conceived and supervised the project and acquired funding. All authors reviewed and edited the manuscript and approved the submitted version.

\section*{Competing interests}
The authors declare no competing interests.

\section*{AI use disclosure}
Large language models assisted in drafting candidate benchmark record descriptions under human chemistry review. Claude (Anthropic), Codex (OpenAI) and Gemini (Google) assisted with literature organisation, analysis scripts, figure preparation and manuscript editing. The authors verified all benchmark records, analyses, citations and scientific interpretations and take full responsibility for the content of the manuscript. {}
\section*{References}
\providecommand{\newblock}{}

\end{document}